\documentclass{article}
\usepackage{graphicx}
\usepackage{xcolor}
\usepackage{mathtools}
\usepackage{geometry}
\usepackage{array}
\usepackage{amssymb}

\usepackage{moreverb,url}
\usepackage{multirow}
\usepackage{caption}
\usepackage{subcaption}
\usepackage{soul}

\newcommand\BibTeX{{\rmfamily B\kern-.05em \textsc{i\kern-.025em b}\kern-.08em
T\kern-.1667em\lower.7ex\hbox{E}\kern-.125emX}}

\title{A Simulation Study of Passing Drivers' Responses to the Autonomous Truck-Mounted Attenuator System in Road Maintenance}
\author{%
    \textbf{Yu Li}\\
  Department of Civil Engineering\\
  Stony Brook University, Stony Brook, NY 11794, USA\\
  Email: yu.li.5@stonybrook.edu\\
  ORCID: 0000-0002-7245-0284\\
  \textbf{Bill Wang}\\
  Department of Civil Engineering\\
  Stony Brook University, Stony Brook, NY 11794, USA\\
  Email: billandoel123@gmail.com\\
  \textbf{William Li}\\
  Department of Civil Engineering\\
  Stony Brook University, Stony Brook, NY 11794, USA\\
  Email: william.li@stonybrook.edu\\
  \textbf{Ruwen Qin}\\
  Department of Civil Engineering\\
  Stony Brook University, Stony Brook, NY 11794, USA\\
  Email: ruwen.qin@stonybrook.edu\\
  ORCID: 0000-0003-2656-8705
}
\date{}
\begin{document}
\maketitle

\begin{abstract}
The Autonomous Truck-Mounted Attenuator (ATMA) system is a lead-follower vehicle system based on autonomous driving and connected vehicle technologies. The lead truck performs maintenance tasks on the road, and the unmanned follower truck alerts passing vehicles about the moving work zone and protects workers and the equipment. While the ATMA has been under testing by transportation maintenance and operations agencies recently, a simulator-based testing capability is a supplement, especially if human subjects are involved. This paper aims to discover how passing drivers perceive, understand, and react to the ATMA system in road maintenance. With the driving simulator developed for this ATMA study, the paper performed a simulation study wherein a screen-based eye tracker collected sixteen subjects' gaze points and pupil diameters. Data analysis evidenced the change in subjects' visual attention patterns while passing the ATMA. On average, the ATMA starts to attract subjects' attention from 500 ft behind the follower truck. Most (87.50\%) understood the follower truck's protection purpose, and many (60\%) reasoned the association between the two trucks. Nevertheless, nearly half of the participants (43.75\%) did not recognize that ATMA is a connected autonomous vehicle system. While all subjects safely changed lanes and attempted to pass the slow-moving ATMA, their inadequate understanding of the ATMA is a potential risk, like cutting into the ATAM. Results implied that transportation maintenance and operations agencies should consider this in establishing the deployment guidance.
\end{abstract}

\section{Introduction}
Roadway transportation is a dominating mode as approximately 82.6\% \cite{mcguckin2018summary} of trips in the U.S. use private vehicles. The U.S. has over 4 million miles of public roadways that need workers and equipment for maintenance, preservation, and rehabilitation. Work zones are hazardous for both nearby traffic and workers there. The work zone fatalities in the U.S. increased to 857 in 2020 \cite{FHWAworkzone}. Compared to a stationary work zone, more factors would influence the safety of a moving work zone, such as the work zone moving speed, the position on the road, and the traffic condition \cite{gan2021evaluation}. The Truck-Mounted Attenuator (TMA) has been adopted to enhance the visibility of work zones to passing vehicles, inform them to change to a work-free lane, and protect workers and equipment at work zones from rear-end crashes \cite{mosunmola2022countermeasures}. TMAs have made positive contributions to safety enhancement, saving about \$23,000 per crash \cite{humphreys1991guidelines}. 23 States that adopted TMAs believe that TMAs are very effective \cite{wang2013evaluation}. However, drivers of shadow vehicles are still likely to get hurt because they stress about being hit and are inevitably in danger of collisions.

With the rapid development of autonomous driving technology, a novel system named Autonomous Truck-Mounted Attenuator (ATMA) is developed by removing the shadow vehicle's driver. The ATMA system includes a lead truck and an unmanned follower truck mounted with an attenuator. The lead truck performs tasks on the road, such as striping, sweeping, and bridge flushing. The unmanned follower truck duplicates the movement of the lead truck and protects it from behind. Unmanned follower trucks have been deployed since 2015 \cite{Driverless_TMA, ATMA_Royal, ATMA_Kratos}. The ATMA system has been under testing or implementation by transportation maintenance and operations agencies in multiple States such as Colorado \cite{ATMA_CO}, Minnesota \cite{ATMA_MN}, Tennessee \cite{ATMA_TN}, Virginia \cite{ATMA_VG}, Missouri \cite{tang2021evaluation}, and others.  While promising test results suggest that the ATMA system can function as expected and its performance is acceptable, the AMTA system is still an emerging technology at the current stage. Detailed operational planning and guidance are needed.

Hence, Tang et al. worked on the operational design domain problem \cite{tang2021identification} and later developed operation guidelines \cite{tang2022development} for the ATMA system. Besides, workers' perception of the ATMA's effectiveness in work zones was also evaluated \cite{pourfalatoun2021user}. Not much work has studied drivers passing the ATMA system. The development of ATMA operation guidance will benefit from understanding the behavior of drivers in the surrounding traffic.

There are three major concerns with drivers in the surrounding traffic. Firstly, they may fail to change the lane timely due to the large speed difference between the ATMA system (from 5 to 15 mph) and the passing traffic. Secondly, vehicles passing the ATMA are not supposed to cut into the ATMA system. Improper understanding of the ATMA system by drivers in the surrounding traffic is a safety concern. Thirdly, drivers may fail to recognize the connected vehicles and autonomous driving technologies implemented in ATMA. It also needs to be determined whether the correct recognition will impact the behavior of drivers in the surrounding traffic. In support of the further development of ATMA operation guidelines, a study is needed to understand how drivers visually perceive, understand, and react to the ATMA system mixed with the traffic.

Driving simulators have been an essential tool for automotive human factors research, such as driving behavior \cite{irwin2017effects}, autonomous systems development \cite{ulahannan2020user}, and road design \cite{filtness2017safety}. Driver simulators are safer, more cost-effective, and easier than field studies for collecting related data. According to specifications of a research topic, simulated driving scenarios are created. Some driving simulators have been designed for TMA studies \cite{bham2010younger,qing2019evaluation,zhang2019simulator}, but not for the ATMA.

The study in this paper fills some of the identified gaps from three aspects. First, it designs and develops a driving simulator for ATMA studies and makes it available to the research community. Then, it develops methods to collect data using the simulator and other tools, as well as defines measurements for analyzing drivers' visual perception, understanding, and reaction when approaching and passing an ATMA system. Additionally, this study has findings that support the further development of ATMA operation guidelines. The remainder of the paper is organized as the following. The next section will review the related literature, followed by the presentation of the research methodology. After that, collected data will be analyzed to discover new knowledge about drivers passing an ATMA system. In the end, important findings and future work will be summarized.

\section{Literature Review}

The literature of this study are in three streams: the ATMA research, relevant simulator-based studies, and analyses of driving behavior using an eye tracker. The literature is briefly discussed below.

\subsection{ATMA Studies}

Most literature on the ATMA system is technical reports from different State Departments of Transportation (DOTs) \cite{ATMA_MO, ATMA_MN, ATMA_TN, ATMA_VG}. The tests were from various perspectives, including communication, the following distance and accuracy, obstacle detection, emergency situations, and more. Among the few academic research about the ATMA system, most were conducted by Tang et al. \cite{tang2021evaluation,tang2021identification,tang2022development}. Their first study proposed to evaluate the system’s performance using statistical models and hypothesis tests \cite{tang2021evaluation}. Results suggested that the ATMA system can function consistently, and the performance is acceptable, stable, and repeatable. Later, the ATMA system is deployed rapidly by multiple state DOTs. Due to the lack of guidance, regulations, and standards, State DOTs deployed the ATMA system using their own criteria of the annual average daily traffic (e.g., a threshold of 6,000 in Colorado DOT). To determine the reasonable traffic condition for the ATMA deployment, Tang et al. employed microscopic traffic flow models \cite{tang2021identification} with the total delay and traffic density as the measurements to evaluate the impact of the ATMA to the traffic flow. Their results support the identification of ATMA's operational design domain. Tang et al. \cite{tang2022development} modeled the driving behavior of the ATMA vehicles at critical decision-making locations by investigating three more technical requirements: the car-following distance, critical lane-changing gap distance, and intersection clearance time. The modeling output suggested important thresholds to use. 
Besides the technical research on the ATMA system, evaluating work zone workers' perception of the usefulness and capability of the ATMA system is also critical to the investment and deployment of the ATMA in work zones. To evaluate user acceptance of the ATMA technology, a survey study was conducted with 13 DOT workers who had some experiences with ATMA systems either in real-world applications or training \cite{pourfalatoun2021user}. Survey results indicate the overall positive acceptance and concerns under various working contexts. Survey results also imply that more training and experience will build a greater trust in the ATMA system among workers.

\subsection{Simulator Studies Related to ATMA}
Driving simulators have been widely adopted to study driver behavior, vehicle design, and road design \cite{irwin2017effects,ulahannan2020user,filtness2017safety}. While no simulator has been developed and widely used for the ATMA system, some simulators were created for studying the TMA system \cite{bham2010younger,qing2019evaluation,zhang2019simulator}. These studies were mainly focused on evaluating TMA's markings and warning signs by designing alternatives. As early as 2010, Bham et al. \cite{bham2010younger} developed a driving simulator to evaluate the effectiveness of four truck-mounted attenuator markings at work zones. These are a lime green and black inverted `V' pattern, a red and white checkerboard pattern, a yellow and black inverted `V' pattern, and an orange and white vertical striped pattern. 
Qing et al. \cite{qing2019evaluation} tested traffic control methods at work zones using ZouSim Driving Simulator.  In this evaluation, drivers' approaching speed and the full-stop distance in two scenarios were analyzed and compared. One scenario uses a  human flagger, and the other uses an automated flagger assistance device on a TMA. Zhang et al. \cite{zhang2019simulator} in the same research team used the same simulator to evaluate four light-color configurations (i.e., amber/white, green only, green/amber, and green/white) for TMAs to improve upon the traditional amber/white lights. 

\subsection{Eye trackers for Studying Driving Behavior}
Eye trackers are used for understanding human visual attention by collecting and analyzing gaze data. It has been adopted by transportation safety studies for more than two decades, especially for evaluating training programs' effectiveness for novice and older drivers \cite{fisher2007empirical}. The evaluation of light-color configurations by Zhang et al.\cite{zhang2019simulator} also employed an eye tracker to investigate the possible effect of disability glare. Generally, the metrics used in eye tracking studies include the gaze distribution \cite{fisher2007empirical,zhang2019simulator,suh2006relationship,louw2017you} for locating where people look at, the fixation duration\cite{jang2014driver} for attention measurement, and pupil parameters\cite{darshana2014efficient, jang2014driver,peysakhovich2015pupil,appel2018cross} for indicating the cognitive load.

\section{Methodology}
This section introduces the simulator design and experiments, the data collection method, and measures defined for this study. 

\subsection{Simulator Design and Development}

The driving simulator is created in Unity 3D. Participants interact with the simulated driving scene through a wheel-pedals set. The design of the simulator is focused on the major elements that are discussed below.

\subsubsection{The Road.}
\begin{figure*}
    \centering
   \includegraphics[width=1\columnwidth]{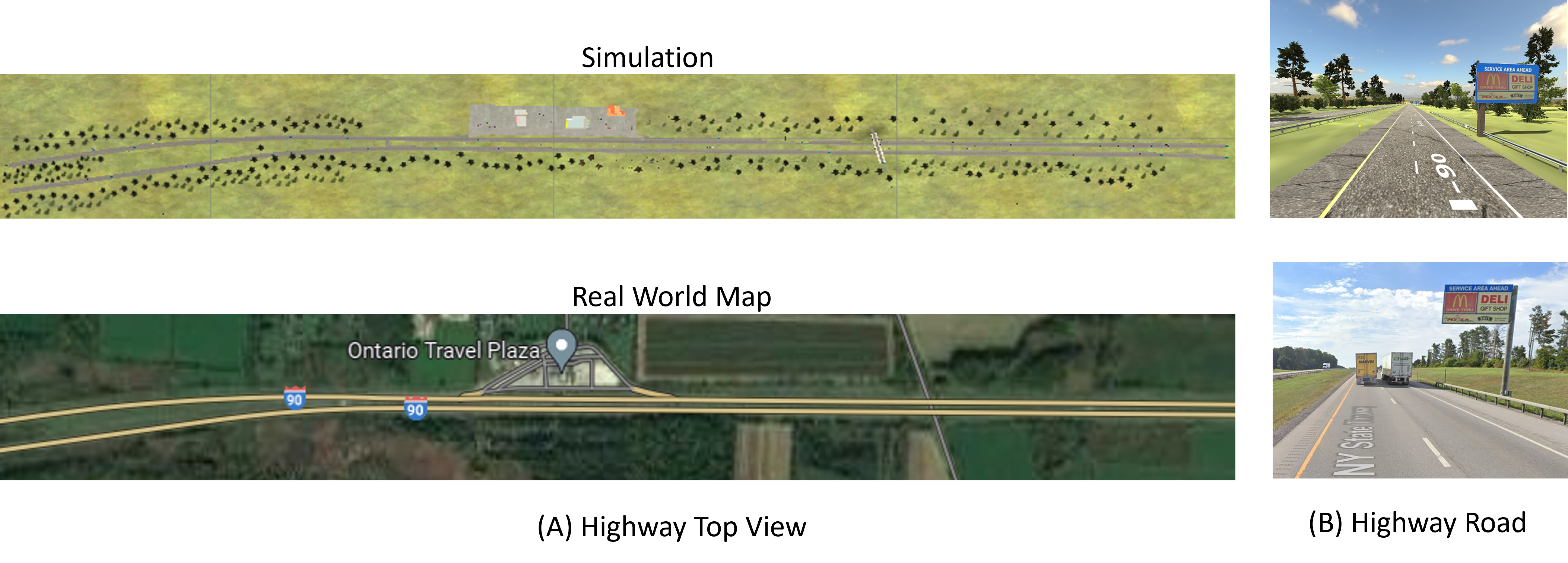}
    \caption{The driving scenario}
    \label{fig:Driving_Scenario}
\end{figure*}

The simulator uses a Unity asset named Road Architect \cite{RoadArchitect} to create the highway road. To mimic the real-world context, the simulator uses the Google Map API to add real road information and geo data to Unity, creating a 2.2-mile divided highway with two lanes in each direction. The road segment is selected from the interstate highway I-90 in the State of New York around the GPS coordinates of (43.030262, -77.906647). The speed limit of this segment is 65 mph, and it contains a rest area where the driver can leave the highway in the simulation. Figure \ref{fig:Driving_Scenario} illustrates the top view of the entire road segment and one sample of the driving scenes, both from the simulation and in the real world. 

\subsubsection{Views.}
\begin{figure*}
\centering
\includegraphics[width=1\columnwidth]{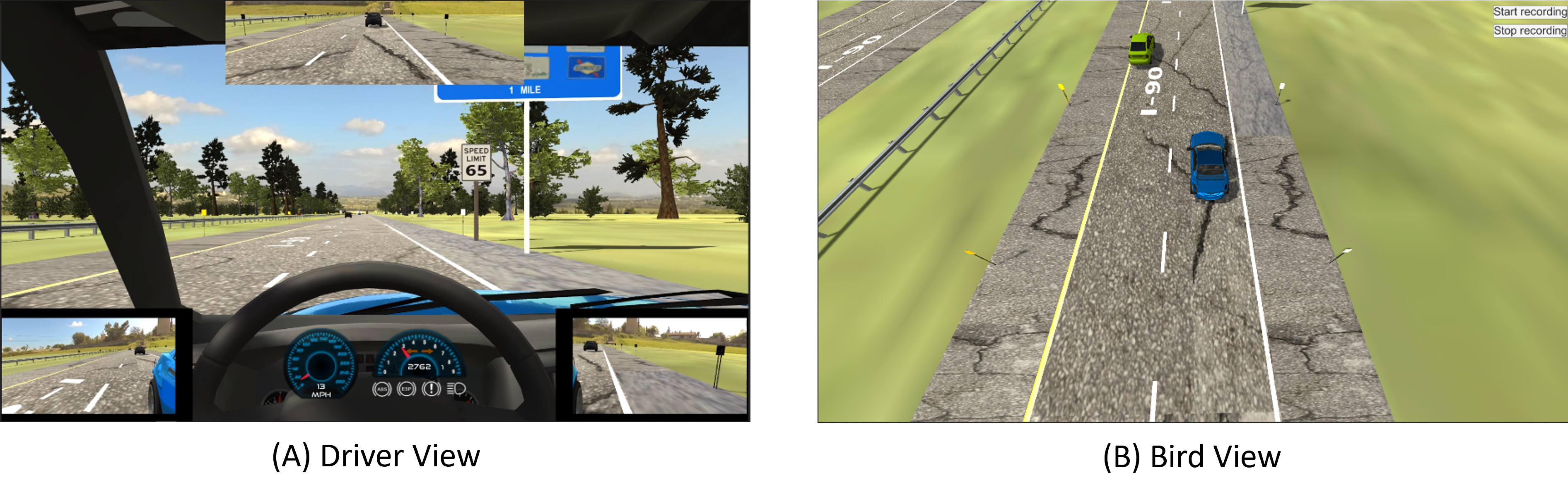}
\caption{Simulator views\label{fig:Driving_views}}
\end{figure*}

The simulator provides participants with the driver's view shown in Figure \ref{fig:Driving_views} (A). It includes the front view, the back views from three mirrors (rear, left, and right), and the dashboard that displays the ego vehicle's speedometer and tachometer with the turn signal indicators on it. In combination with a screen-based eye tracker, the simulation study can determine where and how long a participant is looking at. A bird view shown in Figure \ref{fig:Driving_views} (B) is provided to the experiment coordinator who can observe the experiment in process and control the experiment setting easily from another monitor.

\subsubsection{Vehicles and the Traffic Volume.}
\begin{figure*}[htbp]
    \centering
    \includegraphics[width=1\columnwidth]{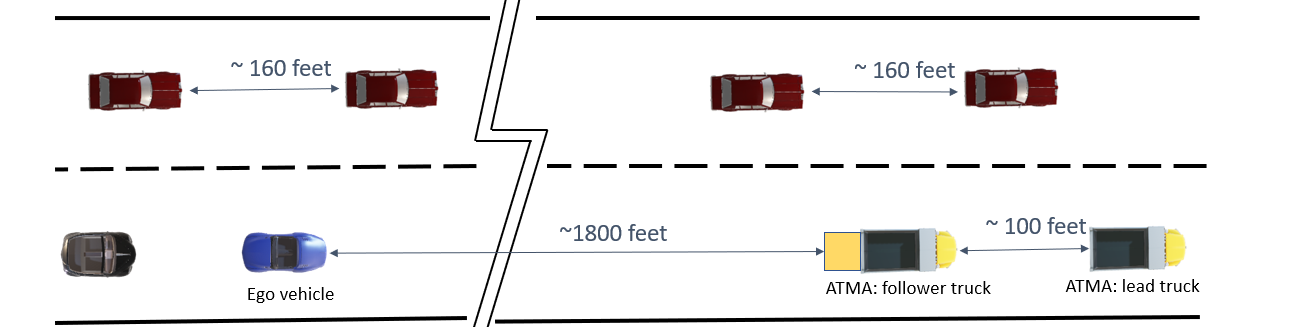}
    \caption{Schematic diagram of the vehicle deployment at the beginning of simulation and in the high traffic volume scenario}
    \label{fig:deploy_high_traffic}
\end{figure*}

Figure \ref{fig:deploy_high_traffic} illustrates the deployment of vehicles and traffic in the simulator. A vehicle model from a Unity realistic car controller asset \cite{RCCUnity} is adopted and revised to become the ego vehicle (blue) that a participant operates in the driving simulation. An ATMA system (yellow) that consists of a follower truck and a lead truck is designed. The ATMA's maximum speed is 15 mph, while the maximum speed for other vehicles is 65 mph. The gap between the lead truck and follower truck is set as 100 ft, approximately. At the beginning of the simulation, the ego vehicle is about 1,800 ft behind the ATMA. Due to the speed difference, the ego vehicle will be approaching and getting closer to the ATMA. All other vehicles are background vehicles, termed Non-Playable Vehicles (NPV). A NPV (black) is behind the ego vehicle, and other NPVs (brown) are mainly on the left lane. Two levels of traffic volume are chosen because the traffic could impact drivers' behavior. The average space between two NPVs on the same lane is about 650 ft in the low traffic volume scenario, and 160 ft in the high traffic volume scenario.

\subsubsection{Interface Between Participants and the Ego Vehicle.}
Participants interact with the ego vehicle in simulation using a Logitech G29 driving force racing wheel and the floor pedals \cite{Logitech_G29}. The ego vehicle has an automatic transmission. Therefore, only a ``Brake'' pedal and an ``Accelerator'' pedal are used in the simulation. Other functions, such as the left and right turn signals, power-off, reversing, and headlight, are also coded for the ego vehicle. Participants can press designated buttons on the wheel to use those functions.

\subsection{Design of Experiments}

This study obtained the institutional IRB's approval, requiring participants to be at least 18 years old with a driver's license. The simulation study for each participant has two experiments, one in the high traffic volume scenario and the other in the low volume scenario. Therefore, the total number of experiments is twice the number of participants.

\begin{figure*}[htbp]
    \centering
    \includegraphics[width=1\columnwidth]{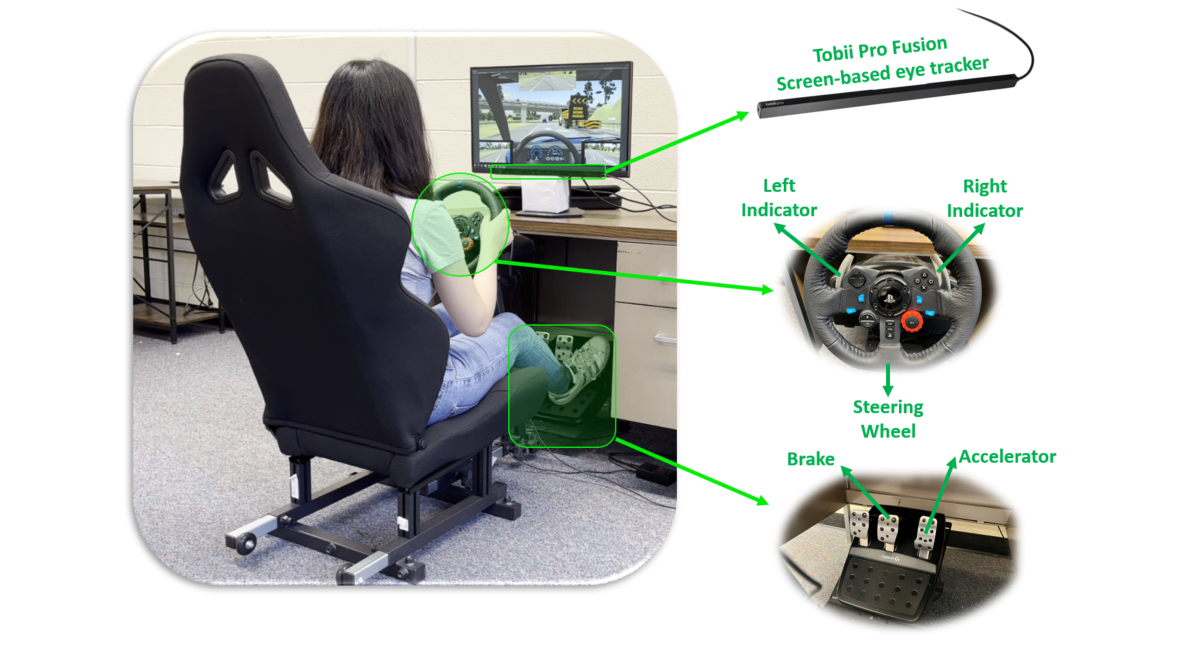}
    \caption{The experiment environment}
    \label{fig:experiment_example}
\end{figure*}

The experiment protocol for this simulation study is the following. In the beginning, an instruction of the driving simulator is presented to the participant. Then, the participant is offered an opportunity to practice the simulator to be familiar with its functions and the virtual driving environment. The driving scene for practice is on the same highway with a lower traffic volume and without the ATMA system. The practice session ends when the participant feels comfortable and confident about operating the driving simulator. To capture participants' perceptions and reactions to ATMA without biases, they are not provided with any information about the ATMA before they perform experiments. The study coordinator describes the driving task as requiring the participant to drive on the highway as usual and take the exit to the next rest area. Before a formal experiment starts, the eye tracker is calibrated for the participant. Then, the participant performs two experiments, one in a low traffic volume scenario and the other in a high volume scenario. Afterward, the participant fills out the questionnaire and exits the study. The entire study for each participant lasts about 30 minutes, while the driving time in each experiment is only about 1$\sim$2 minutes. Figure \ref{fig:experiment_example} illustrates the experiment environment.

\subsection{Data Collection}

Three types of data were collected from the simulator study, and they are presented below.

\subsubsection{Simulation Data Collection.}
The simulator renders 60 frames of driving simulation per second (fps). The streaming data of the simulation are collected frame by frame with an index of $i$. For each experiment, the starting frame is defined as one when the driver is stepping on the accelerator, and the ending frame corresponds to the time when the driver reaches the highway exit. Besides the image frames of the simulation and their timestamp, the simulator can collect three types of data for each frame: operation data, $O_{i}$, position data, $P_{i}$, and display data, $D_{i}$.

The operation data, $O_i$, associated with frame $i$ include the brake input, $o_{b,i}$, the accelerator pedal input, $o_{a,i}$, and the speed of the ego vehicle, $o_{v,i}$:

\begin{equation}
O_{i}=[o_{b,i}, o_{a,i},o_{v,i}],
\end{equation}
where $o_{b,i}$ and $o_{a,i}$ both take values within the range $[0, 1]$, and $o_{v,i}$ is a non-negative continuous variable in the unit of mph. While not used in this specific study, the simulator is designed to record other inputs including the wheel steering, the status of the left/right turn indicators, power-off, reversing, and headlight status collected during the study. Therefore, their notations are skipped in this paper.

\begin{figure*}[htbp]
    \centering
    \includegraphics[width=1\columnwidth]{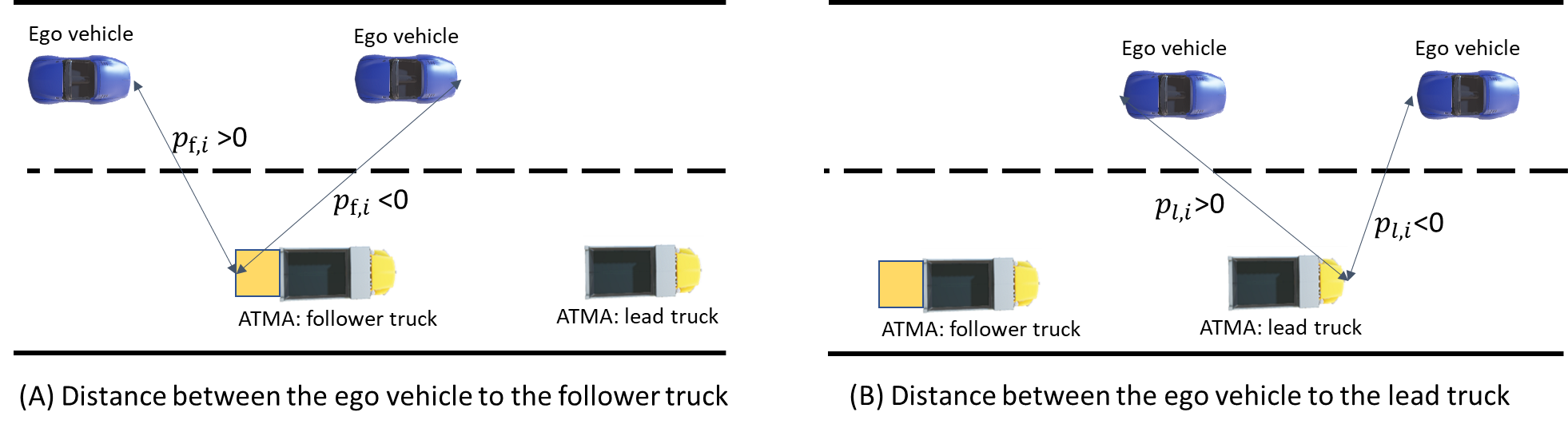}
    \caption{Definition of the lead truck and the follower truck positions relative to the ego vehicle}
    \label{fig:position_Data}
\end{figure*}

The position data associated with frame $i$, $P_i$, include the ego vehicle's position relative to the exit, $p_{e,i}$, the follower truck's position relative to the ego vehicle, $p_{f,i}$, and the lead truck's position relative to the ego vehicle, $p_{l,i}$: 

\begin{equation}
P_{i}=[p_{e,i}, p_{f,i},p_{l,i}],
\end{equation}
where $p_{e,i}$ is non-negative, and the way of defining $p_{f,i}$ and $p_{l,i}$ are illustrated in Figure \ref{fig:position_Data}. $p_{f,i}$ is the distance from the end of the follower truck to the front of the ego vehicle. $p_{f,i}>0$ if the follower truck is in front of the ego vehicle. $p_{l,i}$ is the distance from the front of the lead truck to the end of the ego vehicle. $p_{l,i}>0$ if the lead truck is in front of the ego vehicle. 

\begin{figure*}[htbp]
    \centering
    \includegraphics[width=1\columnwidth]{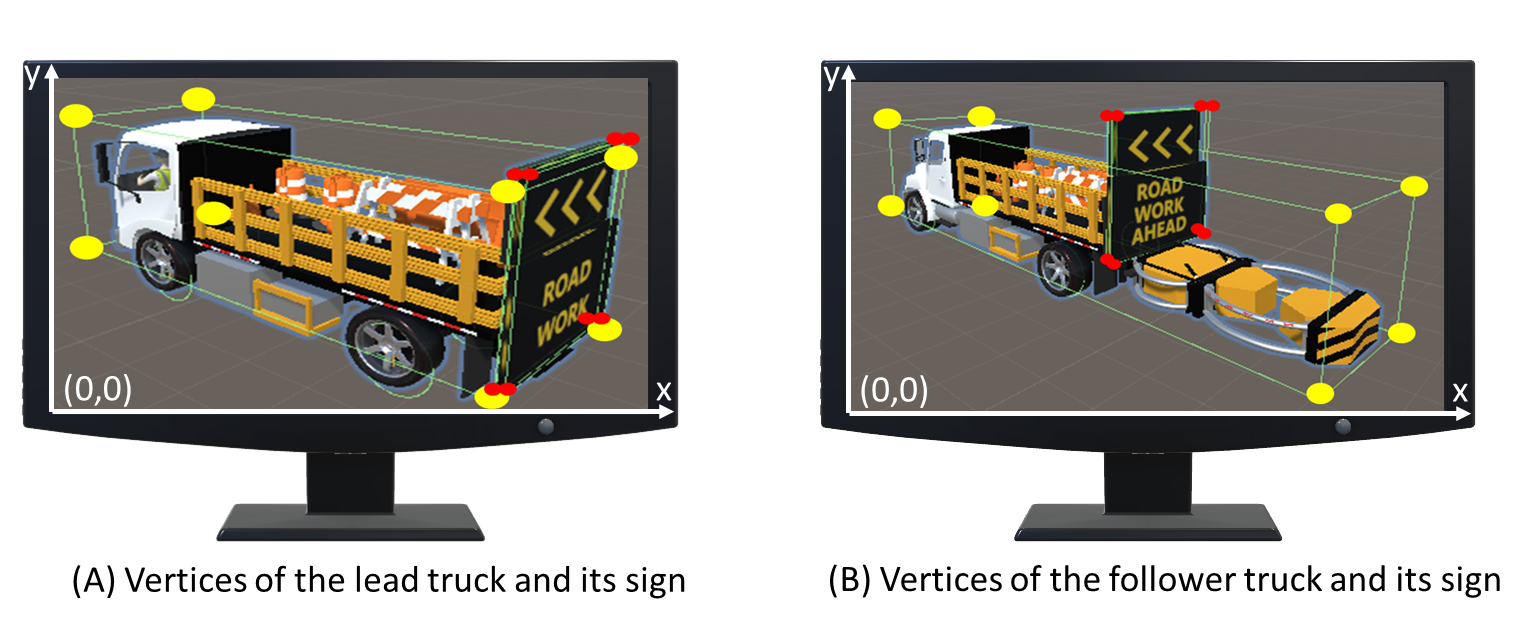}
    \caption{Illustration of object vertices on the screen coordinate system  }
    \label{fig:8vertex}
\end{figure*}

To determine the position and size of each truck or its sign in a frame, the smallest cuboid containing the object is defined, as Figure \ref{fig:8vertex} illustrates. After projecting the eight vertices of the cuboid onto the display screen, their coordinates in the frame are collected:

\begin{equation}
    D_i=[D_{f,i}, D_{l,i}, D_{fs,i}, D_{ls,i}]
\end{equation}
where $D_{f,i}=\{(d_{fx1,i},d_{fy1,i}), \dots (d_{fx8,i},d_{fy8,i})\}$ are the coordinates of the eight vertices defining the follower truck. $D_{l,i}$, $D_{fs,i}$, and $D_{ls,i}$ define the lead truck, the sign of the follower truck, and the sign of the lead truck, respectively. In Figure \ref{fig:8vertex}, the yellow dots in a frame are projected vertices for defining a truck, and the red dots represent those for defining the sign of the truck.

\subsubsection{Gaze Data Collection.}

The simulation study used a screen-based eye tracker, Tobii Pro Fusion \cite{tobiiprofusion}, and the software, Tobii Pro Lab, to capture and record each participant's gaze point coordinates, $G_i=[g_{x,i}, g_{y,i}]$, fixation coordinates, $F_i=[f_{x,i}, f_{y,i}]$, and the pupil diameters of both eyes, $U_i=[u_{z,i}, u_{r,i}]$, when watching any frame $i$. To keep consistent with the simulation data, the frequency of the eye tracker is set as 60 Hz. According to the angular speed of gaze points, $G_{i}$ is classified as a fixation point, a saccade point, or an unclassified point using the Tobii I-VT fixation filter \cite{olsen2012tobii}. A group of successive fixation points of a participant is termed a fixation. The coordinates of the fixation are at the center of the fixation points belonging to this fixation. It should be noted that the gaze point coordinates $G_i$ can change from one frame to another, but the fixation coordinates $F_i$ may keep unchanged in a series of successive frames.

\subsubsection{Questionnaire Data Collection.}
A questionnaire is designed to have three sections. The first section asks about participants' age, gender, and driving experience in mileage per year and in years. Answers to these questions are supplemental information helping better understand participants' driving behavior in the study.

Questions in the second section are listed in Table \ref{tab:drivbehor_questions}, which let participants describe their driving experience in the simulation study. The first question attempts to determine if the ATMA system captured participants' attention in the simulation. The second question in this section is to check whether participants notice the 65 mph speed limit sign at the beginning of the simulation, as shown in Figure \ref{fig:Driving_views} (A). The third question is to collect their judgment of speeding. The last question wants to determine whether the ATMA system is a reason for lane changes.

\begin{table*}[htbp]
\centering
\caption{Questionnaire - Participants' Recall of Driving Experience in the Simulation Study}   
\label{tab:drivbehor_questions}
\small
\begin{tabular}{ m{2cm} m{3cm}  m{2.5cm}  m{3cm}  m{2.5cm} }
\hline\hline
\multicolumn{5}{l}{Are there any particular vehicle(s) captured your attention? }\\
\multicolumn{5}{l}{If there are any vehicle(s) captured your attention, please describe the vehicle in brief. \underline{\hbox to 2cm{}}} \\
\hline\hline
\multicolumn{5}{l}{What is the speed limit of the highway in the experiment?} \\
$\circ$ 55 & $\circ$ 60 & $\circ$ 65 & $\circ$ 70  & $\circ$ I don't know.  \\ 
\hline\hline
\multicolumn{5}{l}{Did you think you were speeding in the experiment?} \\

$\circ$ Yes & $\circ$ Highly likely  & $\circ$ Possible  & $\circ$ Highly unlikely  & $\circ$ No  \\
\hline\hline
\multicolumn{5}{l}{If you changed lane, why did you change lane in the experiment? (select all that applies)} \\

\multicolumn{5}{l}{$\square$ Just for fun.} \\
\multicolumn{5}{l}{$\square$  The vehicle behind me is too close.} \\
\multicolumn{5}{l}{$\square$  The front vehicle is too big.} \\
\multicolumn{5}{l}{$\square$  The sign on the truck(s) suggested lane change.} \\
\multicolumn{5}{l}{$\square$  I need to exit the highway.} \\
\multicolumn{5}{l}{$\square$  The front vehicle is too slow.} \\
\multicolumn{5}{l}{$\square$  Other:  \underline{\hbox to 2cm{}}  } \\
\hline\hline
\end{tabular}
\end{table*}

\begin{figure*}[htbp]
    \centering
  \includegraphics[width=1\columnwidth]{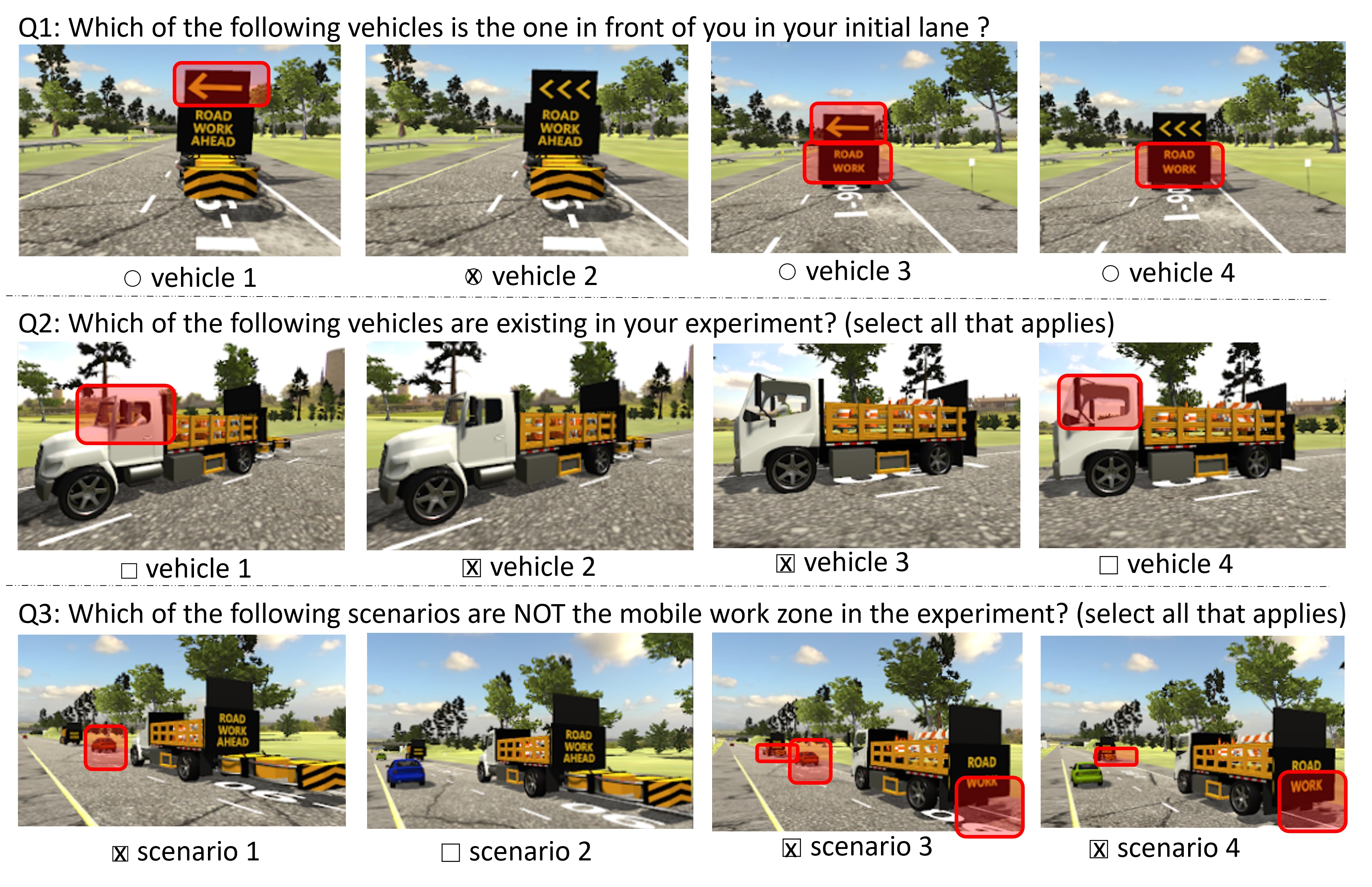}
    \caption{Questionnaire - ATMA Understanding (checked choices are correct answers, red boxes indicate mistakes of wrong choices)}
    \label{fig:atma_questions}
\end{figure*}

The third section includes three questions related to the ATMA system, summarized in Figure \ref{fig:atma_questions}. The first question attempts to determine if the sign and the attenuator mounted on the follower truck captured participants' attention. Vehicles 3 and 4 are the wrong choices as there is no attenuator mounted to the truck.  Vehicles 1 and 3 are wrong because of the incorrect arrow sign. The second question is to find out if participants realize that the follower truck is unmanned. Vehicle 1 is the follower truck supposed to be unmanned. And vehicle 4 is the lead truck that should be driven by a driver. The third question is designed to determine whether participants understand that the two trucks form an ATMA system and that cutting into the system is prohibited. Scenarios 1 and 3 are not a mobile work zone because there is a cut-in. The sequence of the lead truck and the follower truck is wrong in scenarios 3 and 4.

\subsection{Measures}

A set of measures are further defined and calculated using the collected raw data for analyzing participants' response when passing the ATMA system.

\subsubsection{Trimming and Alignment of Multi-modal Time Series Data.}

First, the simulation data collected from each experiment is trimmed by defining the starting and ending frames. The starting frame of each experiment is indexed as $i=1$. Therefore, the first frame is when the recorded accelerator input is different from the preceding frame for the first time (i.e., $o_{a,1}\neq o_{a,0}\; \&\; o_{a,0}=o_{a,w}\,,\forall w<0$ ). The ending frame of each experiment is the one wherein the position of the ego vehicle relative to the exit is minimal (i.e., $\arg\min_i \{p_{e,i}\}$). Then, the timestamps of the starting and ending frame are used to trim the gaze data collected from the experiment. Finally, the trimmed gaze data are aligned with the trimmed simulation data, becoming the dataset of the experiment.

\subsubsection{The Number of Brakes.}
When the participant does not touch it, the brake pedal input, $o_{b,i}$, stays at the steady state, $\alpha_b$, a small value close to zero. $o_{b,i}$ increases and shifts away once the participant steps on and presses the brake pedal. $o_{b,i}$ returns back to the steady state after the participant steps off the brake pedal. A brake is a harsh brake if the average pedal input per second exceeds a threshold $\alpha_h$. Therefore, a brake during an experiment, indexed by $j$, is described as:

\begin{equation}
    B_j=[b_j, q_j, h_j],
\end{equation}
where $b_j$ is the starting frame of the $j$th brake, and $q_j$ is the duration of the brake: 

\begin{equation}
    o_{b, b_j-1}\approx \alpha_b\; \&\; o_{b, b_j+n}>>\alpha_b,\; n=0,1,\dots, q_j,
\end{equation}
and $h_j$ is a binary variable indicating if the brake is a harsh brake,

\begin{equation}
    h_j = \pmb{1}\left\{\frac{o_{b,b_j}-o_{b,b_j+q_j}}{q_j/\text{fps}}>\alpha_h\right\}.
\end{equation}
In this study, fps is 60, and $\alpha_h=10.648$ mph/s chosen according to \cite{kamla2019analysing}. The number of brakes and harsh brakes can be counted respectively using recorded $B_j$'s.

\subsubsection{The Number of Lane Changes.}
A lane change can be detected using one or multiple measures, such as the left/right turn indicators, the ego vehicle's vertical rotation, or the velocity perpendicular to the road direction. Because people have diverse driving behavior, using those measures for detecting a lane change does not guarantee 100\% accuracy. Considering that the number of experiments is not large, the study manually annotated the starting and ending frames of each lane change. Therefore, a lane change event, indexed as $k$, can be described as:

\begin{equation}
    C_k=[s_k, e_k]
\end{equation}
where $s_k$ and $e_k$ are the indices of the starting and ending frames of the $k$th lane change. 

If a participant tried to change lanes to pass the ATMA and then changed lanes back after passing the ATMA, the time period between these two changes is defined as the phase for passing the ATMA,

\begin{equation}
    E =[t_s,t_e],
\end{equation}
where $t_s$ and $t_e$ are the starting and ending frames of the passing phase:

\begin{equation}
\begin{aligned}
    &t_s=\max_{s_k}\{s_k|p_{f,s_k}>0\},\\
    &t_e=\min_{e_k}\{e_k|p_{l,e_k}<0\}.
\end{aligned}
\end{equation}

\subsubsection{Correlation of Ego Vehicle's Velocity and Position to ATMA.}

A Pearson's r value for the ego vehicle's velocity, $o_{v,i}$, and the position of the ego vehicle relative to the ATMA system, $(p_{f,i}+p_{l,i})/2$, in a segment defined by its index set $S$ (i.e., $i\in S$), is calculated to determine their correlation $r_{v,p}$ within that segment:

\begin{equation}
r_{v,p} = \frac{\sigma_{v,p}}{\sigma_v\sigma_p}.
\end{equation}
This correlation value is calculated to determine how the position of the ego vehicle relative to the ATMA system impacts its velocity.

\subsubsection{Gaze on Dashboard and Mirror Areas.}

Binary variables are defined to determine whether the participant looked at the mirrors and the  dashboard that are five fixed areas in each frame:

\begin{equation}
A_{i}=[a_{ms,i}, a_{mt,i}, a_{ml,i}, a_{mr,i}, a_{mb,i}]
\end{equation}
where $a_{ms,i}, a_{mt,i}, a_{ml,i}, a_{mr,i}, a_{mb,i}$ indicate if the participant's gaze point falls on the speedometer, tachometer with turn signal indicators, left mirror, right mirror, or rear view mirror, in frame $i$.

\subsubsection{Gaze on ATMA Trucks.}
\begin{figure}[htbp]
    \centering
   \includegraphics[width=\columnwidth]{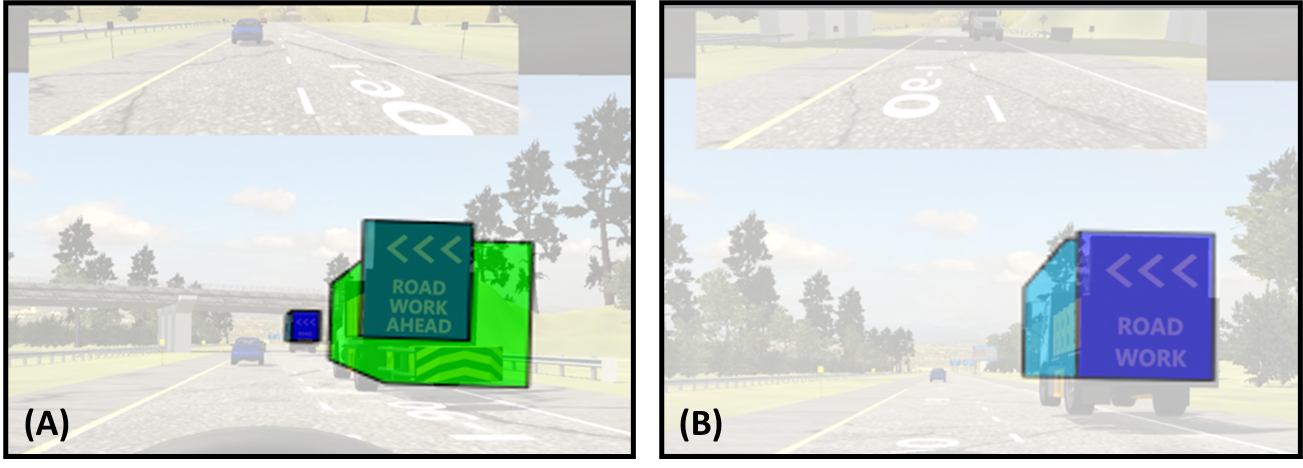}
    \caption{Example Frames with ATMA Masks}
    \label{fig:ATMA_masks}
\end{figure}

Different from the dashboard and mirror areas which are fixed in each frame, the ATMA trucks and their signs are moving from one frame to another. The area in any frame $i$ which contains a moving object is the convex hull containing the object's projection on the frame. Figure \ref{fig:ATMA_masks} shows two frames wherein the convex hulls containing the ATMA components are presented as colored masks covering those components. 

Four binary variables are defined, respectively, to indicate if the participant's gaze falls on the follower truck, lead truck, the sign of the follower truck, and the sign of the lead truck:

\begin{equation}
    M_i=[m_{f,i},m_{l,i},m_{fs,i},m_{ls,i}],
\end{equation}
where $m_{f,i}$, $m_{l,i}$, $m_{fs,i}$, and $m_{ls,i}$ are defined using the spatial relationship between the gaze point, $G_i$, and the convex hull defined for those components, $\{Cov(D_{\gamma,i})|\gamma=f, l, fs, ls\}$:

\begin{equation}
    m_{\gamma,i}=\pmb{1}\{G_i\in Cov(D_{\gamma,i})\},\quad \text{for}\;\gamma=f, l, fs, ls.
\end{equation}

\section{Data Analysis and Results}

The study collected data from 16 participants aged 23 to 45. Their driving experiences are from 1 to 19 years, with median and mean values of 5 and 6.3 years, respectively, and their average annual mileage is 5,000 to 15,000. The gaze data and the simulation data are time series data collected from the 32 experiments (16 subjects $\times$ 2 levels of traffic volume). In total, the collected dataset contains 32 sub-datasets. Each sub-dataset corresponds to one experiment and contains the corresponding gaze data and the simulator data synchronized based on their timestamps. The portion of valid gaze data is around 80\% in three sub-datasets, and it exceeds 90\% in all other sub-datasets. In one of the 32 experiments (participant \# 2 in the high traffic volume scenario), the participant failed to pass the ATMA and changed the lane back to the right to exit the highway behind the ATMA. Those collected and processed data, combined with the questionnaire data, are used to study participants' responses to the ATMA.

\subsection{The Visual Attention to ATMA}

\begin{figure*}[htbp]
    \centering
    \includegraphics[width=1\columnwidth]{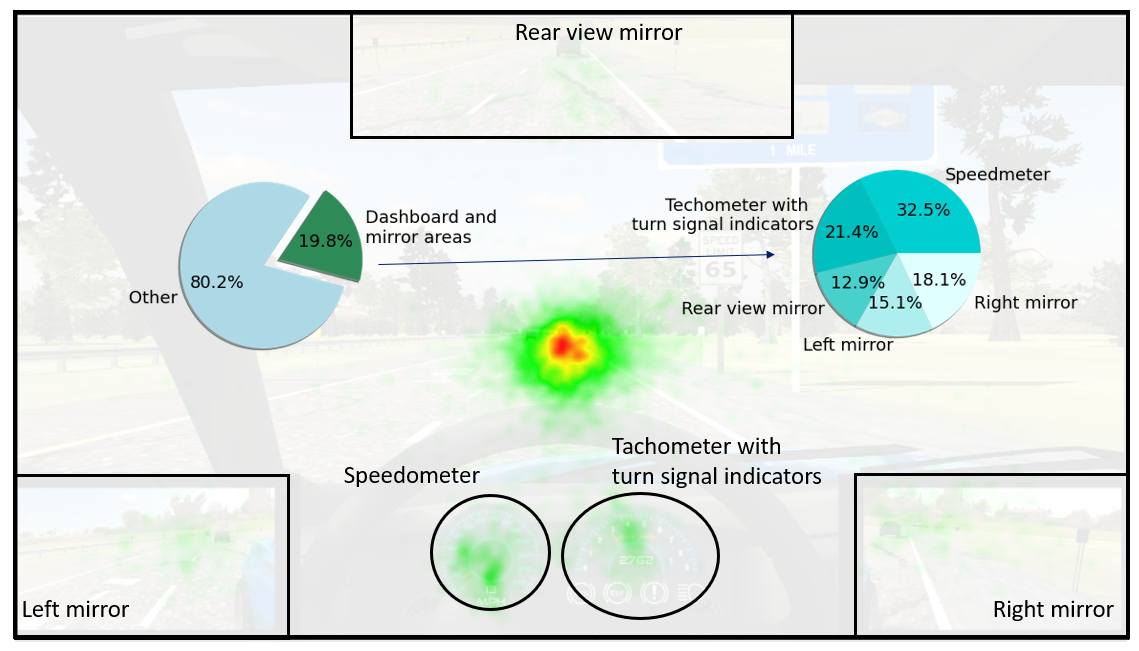}
    \caption{The heat map of fixations}
    \label{fig:AOI}
\end{figure*}

14 participants stated that at least one vehicle caught their attention, and 9 of them mentioned the ATMA in their answers. 
Participants' gaze data are used to analyze their attention related to ATMA. About 138,600 out of 173,000 gaze points (80.1\%) are classified as fixation points. A group of successive fixation points is termed a fixation. In total, the dataset contains 6,527 fixations, and each fixation lasts from 65 to 6,750 milliseconds, with a median duration of 233 milliseconds. The location of a participant's fixation indicates where the participant attends to and the fixation duration indicates the intensiveness of the attention. Figure \ref{fig:AOI} is a heat map that illustrates the spatial distribution of fixation counts. 19.8\% of the fixations are on the mirrors and the dashboard. The remaining 80.2\% are mainly clustered around the front area of the driving scene, forming a hot spot in the heat map.

\subsubsection{Changes in Fixation Distributions When Passing the ATMA.}
\begin{figure*}[htbp]
    \centering
    \includegraphics[width=1\columnwidth]{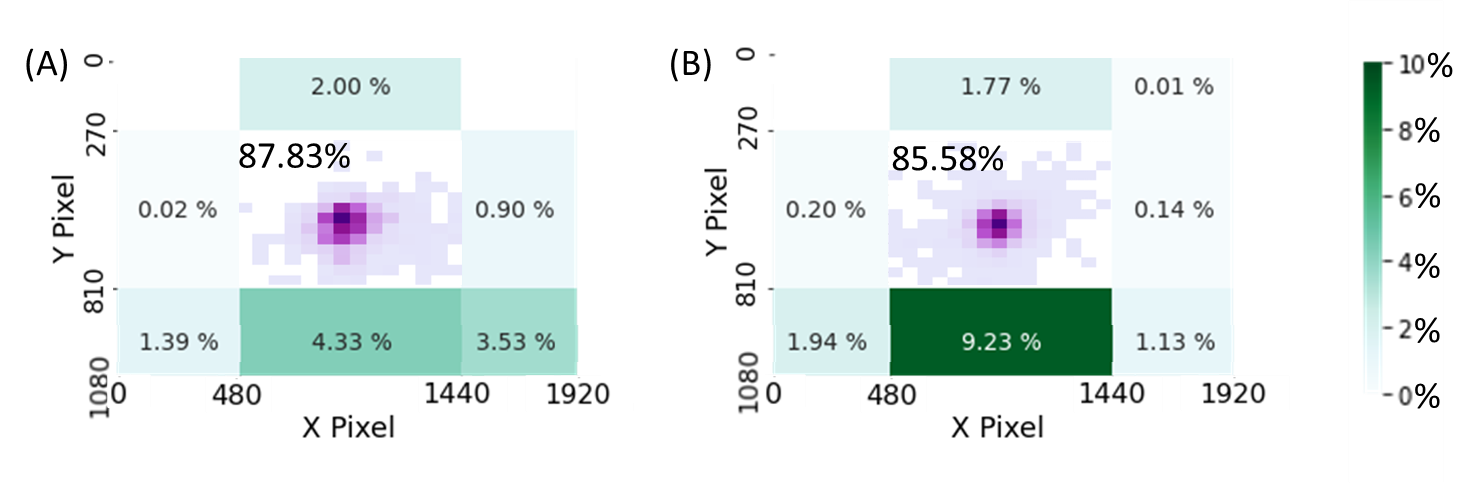}
    \caption{Distribution of fixation durations: (A) passing vs. (B) not passing the ATMA}
    \label{fig:passingATMA}
\end{figure*}

Inspired by the fixation distribution in Figure \ref{fig:AOI}, the screen is split into nine sub-areas. Fixations are split into two mutually exclusive groups according to participants' relative location to the ATMA: when passing the ATMA and otherwise. Figure \ref{fig:passingATMA} shows the distribution of fixation duration for each group. The comparison between the two heat maps indicates changes in the fixation duration. The center is the sub-area that receives the longest attention. During the passing phase, the percentage of cumulative fixation falling in this sub-area is increased by 2.25\% (=87.83\%-85.58\%). If zooming in and examining the detail of this particular sub-area, the distribution of fixation duration within this sub-area has a larger hot spot when passing the ATMA than otherwise. The observations reveal that participants' visual attention to the center sub-area is longer and wider in the passing phase. This may be related to the attention to the follower truck when approaching it from the back. The middle of the bottom is the sub-area containing the dashboard. It is the sub-area that receives the second longest cumulative fixation duration. It is noted that the duration in percent falling in this sub-area drops 4.9\%(=9.23\%-4.33\%), and the reduced amount is shifted to the center, the bottom right, and the center right. One explanation of this phenomenon is that participants were looking at the ATMA on the right of the ego vehicle or appeared in the right mirror.  

\subsubsection{Changes in Pupil Diameters When Passing the ATMA.}
The participants' pupil diameters and their cognitive load are positively correlated. The study split pupil diameters in each sub-dataset into the passing and non-passing phases. Then, it tested the change in each eye's mean pupil diameter for every participant using the two-sample t-test in Table \ref{tab:pupil-t-test}. The test's null hypothesis is that the mean pupil diameter of the same eye in the two phases has no difference. The alternative hypothesis is that the mean pupil diameter in the passing phase is larger than in the non-passing phase. At the level of significance 0.01, all 16 participants have at least one eye that experienced an increase in the mean pupil diameter during the passing phase. The number of participants whose mean pupil diameters of both eyes increased is 9. The test results indicate participants' cognitive load is higher when passing the ATMA, and the increase is likely associated with the increased visual attention to the ATMA during this phase. 

\begin{table*}[htbp]
\caption{P-values of the two-sample t-test of pupil mean diameter: passing vs. not passing the ATMA}   
\label{tab:pupil-t-test}
\footnotesize
\begin{tabular}{m{0.8cm}|m{0.6cm}|m{0.4cm}m{0.4cm}m{0.4cm}m{0.4cm}m{0.4cm}m{0.4cm}m{0.4cm}m{0.4cm}m{0.4cm}m{0.4cm}m{0.4cm}m{0.4cm}m{0.4cm}m{0.4cm}m{0.4cm}m{0.4cm}}
\hline\hline
Traffic & Pupil  &\multicolumn{16}{c}{Subject}  \\
Volume &  & \#1 & \#2& \#3& \#4& \#5& \#6& \#7& \#8& \#9& \#10& \#11& \#12& \#13& \#14& \#15& \#16 \\
\hline\hline
\multirow{2}{1cm}{Low} &Left&	0.00 &	0.00 &	0.00 &	0.00 &	0.00 &	0.00 &	0.00 &	0.00 &	0.00 &	0.00 &	\textcolor{gray}{1.00} &	0.00 &	0.00 &	0.00 &	\textcolor{gray}{0.99} & 0.00 \\
& Right &	0.00 &	0.00 &	0.00 &	\textcolor{gray}{0.12} &	0.00 &	0.00 &	0.00 &	0.00 &  0.00 &	0.00 &	\textcolor{gray}{1.00} &	0.00 &	0.00 &	0.00 &	\textcolor{gray}{1.00} &	0.00 \\
\multirow{2}{1cm}{High} & Left &	0.00 &	0.00 &	-    &	0.00 &	0.00 &	0.00 &	0.00 &	\textcolor{gray}{1.00} &	0.00 &	0.00 &	0.00 &	0.00 &	0.00 &	0.00 &	0.00 &	0.00 \\
& Right & 	0.00 &	0.00 &	0.00 &	0.00 &	\textcolor{gray}{0.06} &	0.00 &	0.01 &	\textcolor{gray}{1.00} &	0.00 &	0.00 &	0.00 &	0.00 &	0.00 &  	- &	0.00 &	0.00 \\
\hline\hline
\multicolumn{18}{l}{ Note: ``-'' means no valid left or right pupil data were collected in the phase of passing the ATMA.}\\
\end{tabular}
\end{table*}

\subsubsection{Fixation on the ATMA.} 
\begin{figure*}[htbp]
    \centering
    \includegraphics[width=1\columnwidth]{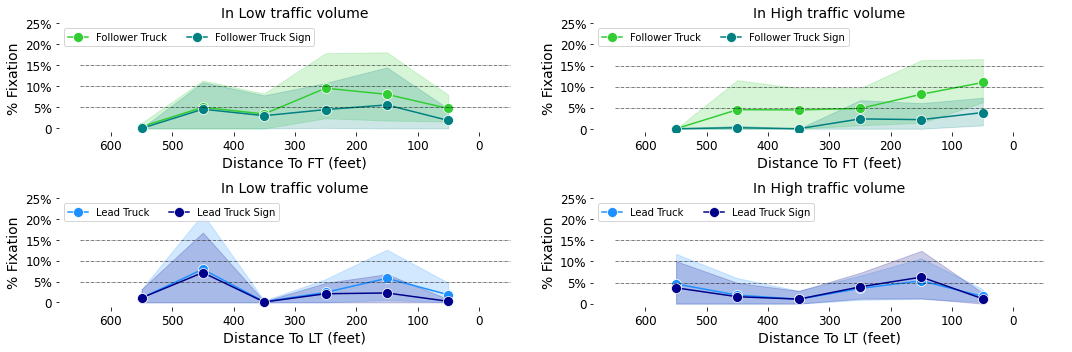}
    \caption{The 95\% interval estimates of fixation points in percent on different components of ATMA}
    \label{fig:fixation_ATMA_distance}
\end{figure*}

A fixation point is identified on the ATMA system if it locates on the follower truck (FT), the lead truck (LT), or their signs. An assumption of this study is that drivers have more attention to the truck that they are approaching. 
In the two upper plots in Figure \ref{fig:fixation_ATMA_distance}, the distance of 600 ft from the back of the follower truck is split into six equally-spaced segments. In each segment, the 95\% interval estimates of a driver's fixation points in percent on the follower truck and on its sign are calculated, respectively. The plot on the upper left is the result for the low traffic volume scenario and the one on the upper right is for the high traffic volume scenario. The two plots on the bottom are similarly defined to study the proportions of fixation points on the lead truck and its sign. 

The two upper plots show that the follower truck and its sign start to attract drivers' attention from 400-500 ft behind. The maximum mean percentage of fixation points on the follower truck reached 10\%, when drivers are 200-300 ft behind the truck in the low traffic volume scenario and 100 ft in the high traffic volume scenario. The maximum mean percentage of fixation points on the sign of the follower truck is 5\%, when drivers are 100-200 ft behind the truck in the low traffic volume scenario. 
The mean percentage of fixation points starts to decrease when the distance to the follower truck is less than 200 ft under the low traffic volume, however it increases under the high traffic volume. Possible reasons for the different patterns observed between the lead truck and the follower truck are the timing of lane change and the speed of the surrounding traffic. As the lane change happened relatively late and the speed was fast in the low traffic volume scenario,  drivers would have fewer opportunities to attend to the follower truck. The earlier lane change and the lower speed in the high traffic volume scenario provide drivers more chances to attend to the follower truck.
The two plots at the bottom do not suggest the same attention behavior when drivers are approaching the lead truck. The lead truck and its sign attract less attention from approaching drivers compared to the follower truck and its sign. This attention pattern seems reasonable because the follower truck and its sign are the ATMA components that first alert approaching drivers about a moving work zone ahead and suggest the lane change. If drivers comprehend that intention, the lead truck and its sign that attempt to keep drivers on the alternative lane are easier for drivers to understand. 

\subsection{Drivers' Understanding of the ATMA}

The study further examines if drivers understand the roles of the follower truck and if they have been aware of the adoption of autonomous driving and connected vehicle technologies for the ATMA system.

\subsubsection{Drivers' Understanding of the Follower Truck's Roles.}
11 out of 16 participants correctly recalled the follower truck and its correct sign in Q1 of Figure \ref{fig:atma_questions}. 3 participants mistakenly chose the answer that the truck has the correct sign but the attenuator is missing. 
That is, 14 participants (87.5\%) can recall the sign on the follower truck, indicating that the sign for alerting drivers to change lanes and pass the ATMA is effective. As 11 participants recall the attenuator mounted on the follower truck but not the lead truck, some drivers probably understand the protection purpose of the follower truck.

\subsubsection{Drivers' Awareness of the Follower Truck Being an Unmanned Vehicle.}

Only half of the participants gave an answer rather than ``I don't know" to Q2 in Figure \ref{fig:atma_questions}. But no participants gave the correct answer. One participant contacted the experiment coordinator a few days after finishing the experiments and shared the experience of seeing a TMA. The participant tried but failed to figure out whether the truck was unmanned due to the low position of the passenger vehicle relative to the truck and the large difference in their speeds. Drivers, particularly those in passenger vehicles, are unlikely to notice that the follower truck is unmanned.

\subsubsection{Drivers' Awareness of the Connected Trucks as an ATMA System.}

\begin{figure*}[htbp]
    \centering
    \includegraphics[width=1\columnwidth]{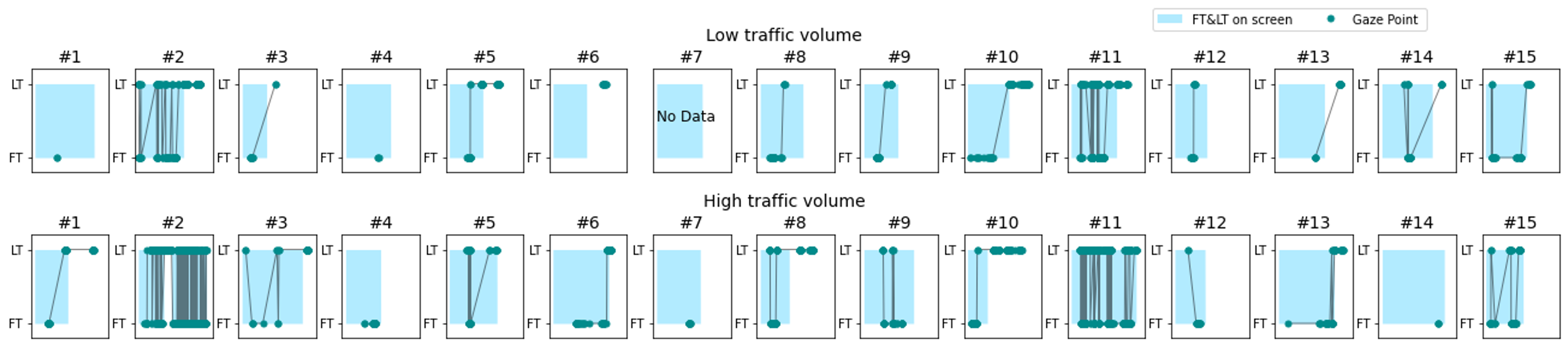}
    \caption{Gaze path on the follower truck (FT) and the lead truck (LT)}
    \label{fig:gaze_ATMA_path}
\end{figure*}

It is rational to infer that a driver does not recognize that the lead truck and the follower truck form an ATMA system if the driver only looked at one truck of the ATMA. The shift pattern of gaze time series, like FT $\rightarrow$ LT $\rightarrow$ FT or LT $\rightarrow$ FT $\rightarrow$ LT, indicates that the driver may attempt to find the association between the two trucks.
Therefore, this shift pattern is used as a necessary condition for inferring drivers' recognition of the ATMA system. 
Figure \ref{fig:gaze_ATMA_path} are line charts of the gaze points falling on the two trucks in the 29 experiments that have valid data for this analysis. The $x$ axis of each chart is the timeline, and the $y$ axis indicates the truck type: the follower truck (FT) and the lead truck (LT).
9 out of 15 participants had at least one gaze shift between the two trucks. That is, 60\% of the participants were reasoning the relationship between the two trucks.

Participants' response to Q3 in Figure \ref{fig:atma_questions} shows 7 of 16 participants (43.75\%) did not recognize that the ATMA is a connected vehicle system. Only one participant (\#14) was explicitly aware of the ATMA as a system and another eight participants gave partially correct answers to this question. Besides participant \#14, answers of another three participants (\#6, \#8, and \#9) indicate they recognized that cutting into the two trucks is prohibited. And 3 of these 4 participants had at least one gaze shift as shown in Figure \ref{fig:gaze_ATMA_path}. 

The observations confirm that drivers' reasoning of the relationship between the two trucks, indicated by the presence of the gaze shift pattern, is a necessary condition for them to recognize that the ATMA is a system when they have no prior knowledge about it. However, their reasoning of the two trucks' relationship in such a short period does not always successfully lead to drivers' understanding of the connected vehicle technology applied to the ATMA.

\subsection{Reactions to the ATMA}
Drivers' reactions to the ATMA system will be discussed from their speed, brake pedal operations, and their lane change behavior. 

\subsubsection{Speed Changes.}

\begin{figure*}[htbp]
    \centering
    \includegraphics[width=1\columnwidth]{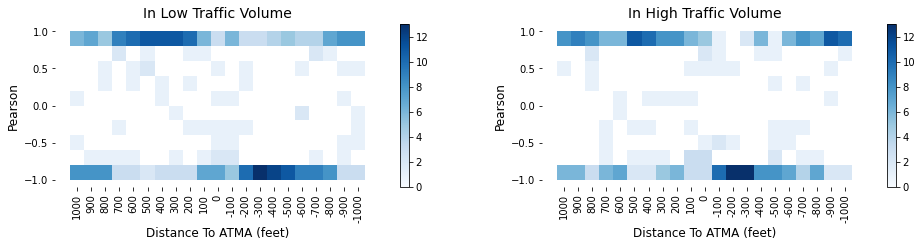}
    \caption{The frequency distribution of Pearson correlations between the ego vehicle's speed and the distance to the ATMA}
    \label{fig:Pearson_speed_ATMA}
\end{figure*}

The hypothesis about the ego vehicle's speed is that the driver will reduce the speed when approaching the follower truck, move with caution when passing along the ATMA, and speed up after passing the lead truck. To verify the hypothesized relationship between the ego vehicle's speed and its distance to the ATMA, the Pearson correlation between the ego vehicle's speed, $o_{v,i}$, and its distance to the center of the ATMA, $(p_{f,i} + p_{l,i})/2$, is calculated for each of the following 21 equally-spaced segments in size 100 ft: 10 segments are behind the follower truck ($p_{f,i}>0$), one segment is from the follower truck to the lead truck, and the rest 10 segments are ahead of the lead truck (i.e., $p_{l,i}<0$). 31 correlation values are calculated for each of the 21 segments (noted that a driver failed to pass the follower truck in one of the 32 experiments). Figure \ref{fig:Pearson_speed_ATMA} are two heat maps showing the frequency distribution of the 651 (=31$\times$21) correlation values. Figure \ref{fig:Pearson_speed_ATMA} clearly shows that the correlation is strongly positive when the ego vehicle is behind the follower truck and switches to strongly negative when the ego vehicle is ahead of the lead truck. For example, in the low traffic volume scenario, the Pearson values of 10 participants are greater than 0.85 when the ego vehicle is 100$\sim$200 ft behind the follower truck, and the Pearson values of 11 participants are less than -0.85 when the ego vehicle is 100$\sim$200 ft ahead of the lead truck. This pattern verifies that participants are likely to reduce their speed when they are approaching and getting closer to the follower truck and speed up after they passed the lead truck. When the ego vehicle is passing the ATMA, no evident pattern of slowing down or speeding up is observed from a major portion of the participants. The observations support the hypothesis about the ego vehicle's speed changes as a reaction to the ATMA. Besides, a switch from the strong negative correlation to the strong positive correlation is noticed when drivers were ahead of the lead truck for 500$\sim$1000 ft in the high traffic volume scenario, but not in the low volume scenario.
The ego-vehicle was closer to the exit when it is 500$\sim$1000 ft ahead of the ATMA in the high traffic volume scenario compared to the low-volume scenario. Despite the heterogenous driving behavior, more participants slowed down when they were close to the exit.

\subsubsection{Brakes.}

\begin{figure*}[htbp]
    \centering
   \includegraphics[width=1\columnwidth]{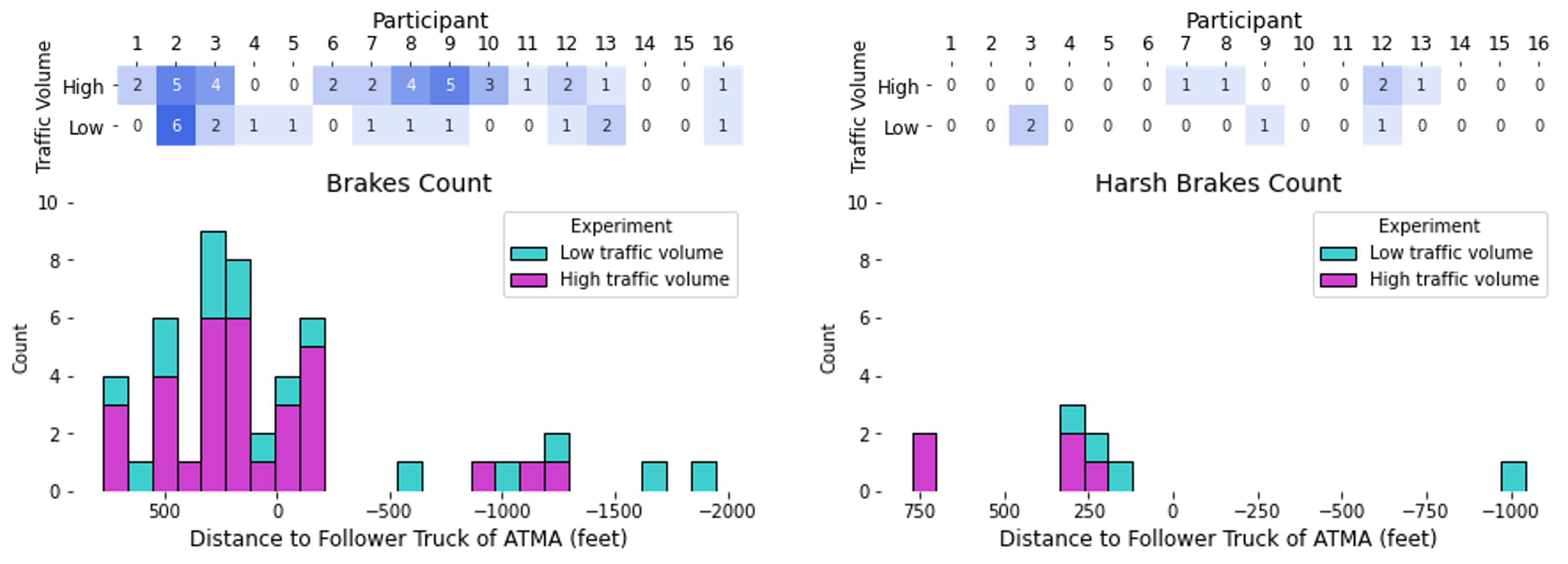}
    \caption{Frequency distribution of brakes and harsh brakes on the distance to the follower truck}
    \label{fig:brakes}
\end{figure*}

Brakes are highly likely when moving towards the slow-moving ATMA and waiting for an opportunity to change lanes, particularly when the traffic volume is high. 49 brakes were detected in this study. The plot on the left of Figure \ref{fig:brakes} counts brakes by the distance to the follower truck, split by levels of traffic volume. 32 out of 49 brakes (65.3\%) occurred in the scenario of high traffic volume, and 41 brakes (83.7\%) were made when the ego vehicle was either behind the follower truck or within 200 ft in front. 9 out of the 49 (18.4\%) brakes are harsh brakes and the plot on the right of Figure \ref{fig:brakes} counts those harsh brakes by the distance to the follower truck. 6 out of the 9 harsh brakes (66.7\%) were made when the ego vehicle was about 100$\sim$300 ft behind the follower truck. It implies that the ATMA may cause more brakes, especially harsh brakes. Therefore, incautious or aggressive drivers are likely to involve in a crash into the ATMA.

\subsubsection{Lane Changes.}

\begin{figure*}[htbp]
    \centering
    \includegraphics[width=1\columnwidth]{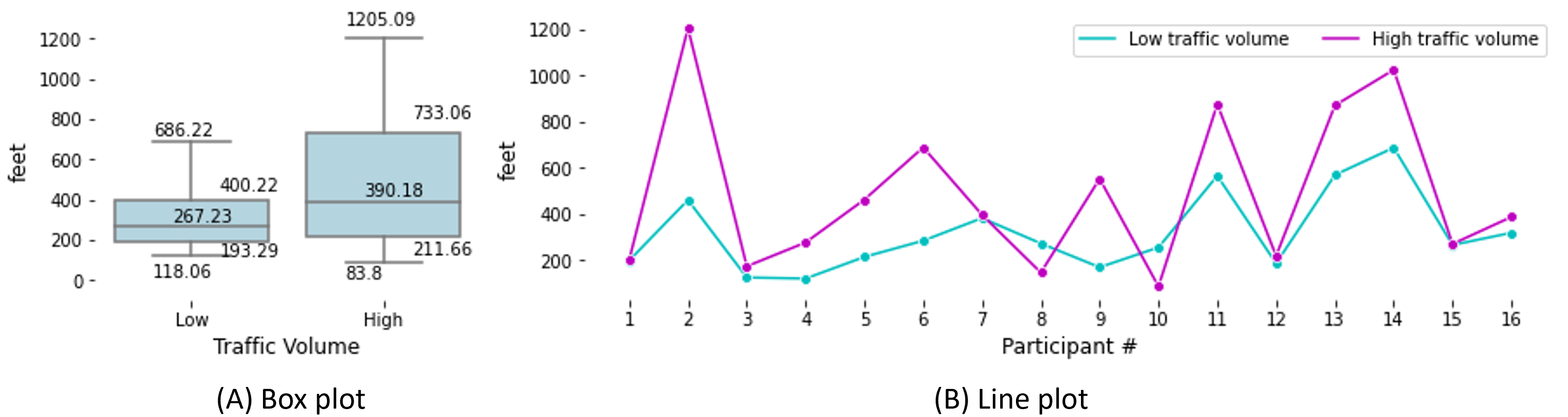}
    \caption{The distance to the follower truck when participants started changing lane to pass the ATMA}
    \label{fig:dis_FT_change_lane}
\end{figure*}

All participants changed lanes during their simulation experiments, and some changed lanes multiple times for various reasons. Figure \ref{fig:dis_FT_change_lane} (B) summarizes the distances of the ego vehicle from the follower truck when the ego vehicle started to change lanes. 14 out of the 16 (87.5\%) participants changed lanes earlier in the high traffic volume scenario to pass the ATMA than in the low traffic volume scenario. Figure \ref{fig:dis_FT_change_lane} (A) further compares the distributions of the distance in the two traffic volume scenarios. The median distance in the low traffic volume scenario is 267 ft and 390 ft in the high traffic volume scenario. 
The interquartile range (IQR) of the distance is 193$\sim$400 ft in the low traffic volume scenario and 212$\sim$733 ft in the high traffic scenario. The comparison reveals that drivers in the low traffic volume scenario were more homogeneously waiting until relatively closer to the follower truck to change lanes, but they tended to change lanes earlier when the traffic volume is high. How earlier they changed the lane varied largely among the participants. The observation indicates drivers may have less pressure about passing the ATMA when the traffic volume is low, but experienced more pressure and thus became more conservative in changing the lane to pass the ATMA.  

\section{Conclusion}

This paper studied drivers’ visual attention, understanding, and reactions when they are passing an ATMA system in road maintenance. A virtual reality-based driving simulator has been designed and developed for this purpose. Methods, metrics, and measures for the data-driven study approach were developed too. Participants' data were collected from the simulation study and analyzed to provide more information for the ATMA operation guidelines from passing drivers’ perspectives. The simulator and experiment data are shared with the public \cite{ATMAdrivingsimulator_Github}.

Important findings from this simulation study support the development of operation guidelines.
First of all, data analysis shows that the nearby traffic is slowing down when approaching and passing the follower truck. This simulation study found that the lane change happened earlier in the high traffic volume scenario than in the low volume scenario, suggesting that the ATMA's impact on the road capacity is dependent on the traffic volume. This finding supports the current practice of operating the ATMA system on low traffic roads. Secondly, the result shows no participant cut into the ATMA in the simulation study although they all have the motivation to change to the right lane and prepare for leaving the highway. However, the result also indicates that 12 out of 16 participants did not explicitly realize that the two trucks form an integrated system and they should not cut into the ATMA when passing it. The gap between the LT and the FT is about 100 ft in our experiment and should be large enough for drivers to cut into the ATAM system. The actual gap could be larger than 100 ft in real practice, like painting lane markings. More effective designs of the warning signals would help drivers in the surrounding traffic better understand that cut-ins are prohibited. Lastly, no participant realizes that the follower truck is unmanned, which indicates the deployment of ATMA to replace TMA would not influence the driving behavior of surrounding traffic. 

The ATMA system's deployment and development should consider passing drivers' knowledge about the ATMA and their reactions when passing it. The analysis in this paper throws the ATMA safety concern from passing drivers due to their inadequate knowledge of the ATMA. 
When the gap between the two trucks reduces, the chance that vehicles in the surrounding traffic will cut into the ATMA system is decreasing. Yet, in some tasks like painting the lane markings, a sufficient gap between the two trucks has to be maintained. The gap between the two trucks is a safety-impacting factor to be examined in detail in a future study.
The simulator and the data-driven method can also be revised for the purpose of training the driver and other operators on the lead truck. 

\bibliographystyle{plain}
\bibliography{Reference}

\end{document}